\documentclass[11pt,a4paper]{article}
\usepackage[hyperref]{acl2017}
\usepackage{times}
\usepackage{latexsym}
\usepackage{amssymb, subfigure, float}
\newfloat{algorithm}{t}{lop}
\usepackage[cmex10]{amsmath}
\usepackage{algorithm}
\usepackage[noend]{algpseudocode}
\usepackage{color, framed, tikz, pgfplots}
\usetikzlibrary{patterns}

\usepackage{url}

\aclfinalcopy 
\def\aclpaperid{9} 

\title{Deep Active Learning for Dialogue Generation}

\author{Nabiha Asghar$^\dagger$, Pascal Poupart$^\dagger$, Xin Jiang$^\ddagger$, Hang Li$^\ddagger$ \\
  $^\dagger$ Cheriton School of Computer Science, University of Waterloo, Canada \\
  {\tt \{nasghar,ppoupart\}@uwaterloo.ca} \\
  $^\ddagger$Noah's Ark Lab, Huawei Technologies, Hong Kong \\
  {\tt \{jiang.xin,hangli.hl\}@huawei.com} \\}

\date{}

\begin{document}
\maketitle
\begin{abstract}
We propose an online, end-to-end, neural generative conversational model for open-domain dialogue. 
It is trained using a unique combination of offline two-phase supervised learning and online human-in-the-loop active learning. 
While most existing research proposes offline supervision or hand-crafted reward functions for online reinforcement, we devise a novel interactive learning mechanism based on hamming-diverse beam search for response generation and one-character user-feedback at each step. Experiments show that our model inherently promotes the generation of semantically relevant and interesting responses, and can be used to train agents with customized personas, moods and conversational styles.
\end{abstract}

\section{Introduction}

\label{intro}
Several recent works propose neural generative conversational agents (CAs) for open-domain and task-oriented dialogue \cite{shang2015neural,sordoni2015neural,vinyals2015neural,serban2015building,serban2016multiresolution,wen2016network,shen2017conditional,eric2017copy,eric2017key}. These models typically use LSTM encoder-decoder architectures (e.g. the sequence-to-sequence (Seq2Seq) framework  \cite{sutskever2014sequence}), which are linguistically robust but can often generate short, dull and inconsistent responses \cite{serban2015building,li2016diversity}. Researchers are now exploring Deep Reinforcement Learning (DRL) to address the hard problems of NLU and NLG in dialogue generation. In most of the existing works, the reward function is hand-crafted, and is either specific to the task to be completed, or is based on a few desirable developer-defined conversational properties. 

In this work we demonstrate how online Deep Active Learning can be integrated with standard neural network based dialogue systems to enhance their open-domain conversational skills. The architectural backbone of our model is the Seq2Seq framework, which initially undergoes offline supervised learning on two different types of conversational datasets. We then initiate an online active learning phase to interact with human users for incremental model improvement, where a unique single-character\footnote{The user has the option to provide longer feedback.} user-feedback mechanism is used as a form of reinforcement at each turn in the dialogue. 
The intuition is to rely on this all-encompassing human-centric `reinforcement' mechanism, instead of defining hand-crafted reward functions that individually try to capture each of the many subtle conversational properties. This mechanism inherently promotes interesting and relevant responses by relying on the humans' far superior conversational prowess.


\section{Related Work \& Contributions}
\label{relatedwork}
DRL-based dialogue generation is a relatively new research paradigm that is most relevant to our work. For task-specific dialogue \cite{su2016continuously,zhao2016towards,cuayahuitl2016deep,williams2016end,li2017investigation,li2017end,peng2017composite}, the reward function is usually based on task completion rate, and thus is easy to define. For the much harder problem of open-domain dialogue generation \cite{li2016deep,yu2016strategy,weston2016dialog}, hand-crafted reward functions are used to capture desirable conversation properties. Li \textit{et al.} \shortcite{li2016simple} propose DRL-based diversity-promoting Beam Search \cite{koehn2003statistical} for response generation. 

Very recently, new approaches have been proposed to incorporate online human feedback into neural conversation models \cite{li2016dialog,abel2017agent,li2017learningthrough}. Our work falls in this line of research, and is distinguished from existing approaches in the following key ways.
\begin{enumerate}
\item We use online deep active learning as a form of reinforcement in a novel way, which eliminates the need for hand-crafted reward criteria. We use a diversity-promoting decoding heuristic \cite{vijayakumar2016diverse} to facilitate this process.
\item Unlike existing CAs, our model can be tuned for one-shot learning. It also eliminates the need to explicitly incorporate coherence, relevance or interestingness in the responses.\\ 
\end{enumerate}

\section{Model Overview}
\label{model}

The architectural backbone of our model is the Seq2Seq framework consisting of one encoder-decoder layer, each containing 300 LSTM units. The end-to-end model training consists of offline supervised learning (SL) with mini-batches of 10, followed by online active learning (AL). 

\subsection{Offline Two-Phase Supervised Learning}
To establish an offline baseline, we train our network sequentially on two datasets, one for generic dialogue, and the other specially curated for short-text conversation.\\ 

\noindent
\textbf{Phase 1}: We use the Cornell Movie Dialogs Corpus \cite{cornellcorpus}, consisting of 300K message-response pairs. Each pair is treated as an input and target sequence during training with the joint cross-entropy (XENT) loss function, which maximizes the likelihood of generating the target sequence given its input.\\  

\noindent
\textbf{Phase 2:} Phase 1 enables our CA to learn the language syntax and semantics reasonably well, but it has difficulty carrying out short-text conversations that are remarkably different from movie conversations. To combat this issue, we curate a dataset from JabberWacky's chatlogs\footnote{http://www.jabberwacky.com/j2conversations. Jabber-Wacky is an in-browser, open-domain, retrieval-based bot.} available online. The network is initialized with the weights obtained in the first phase, and then trained on the JabberWacky dataset (8K pairs). Through this additional SL phase of fine-tuning on a small dataset, we get an improved baseline for open-domain dialogue (Table \ref{SLvsRL}, Figure \ref{learningrate}a).

\def\NoNumber#1{{\def\alglinenumber##1{}\State #1}\addtocounter{ALG@line}{-1}}

\begin{algorithm}[t]
\caption{Online Active Learning}\label{ORL}
  \small
\begin{algorithmic}[1]
\Procedure{HammingDBS(text)}{}
\State \textit{r} = \textit{emptyList(size = K)};
\For {$t = 1 \text{ to } T$}
\State \textit{r}[1][$t$] = model.forward(\textit{text, r}[1][1,...,$t-1$]);
\For {$i = 2 \text{ to } K$} \hfill// \textit{K = 5 in our setting}
\State \textit{augmentedProbs} = model.forward(\textit{t,text,r}[i]) 
\NoNumber{\hspace{0.8cm}$+\lambda$(\textit{hammingDist}($r$[$i$], $r$[1, ..., $i-1$]));}
\State \textit{r}[i][$t$] = top1(\textit{augmentedProbs});

\EndFor
\EndFor
\State return \textit{r};
\EndProcedure
\Procedure{OnlineAL()}{}
\State \textit{lr} $\gets$ \textit{0.001}; \hfill// \textit{initial learningRate for Adam}
\While {\textit{true}}		
\State $\textit{usrMsg} \gets \text{io.read(); }$
\State $\textit{responses} \gets \text{HammingDBS}\textit{(usrMsg}\text{);}$
\State $\text{io.write(}\textit{responses}\text{);}$
\State $\textit{feedback} \gets \text{io.read();}$
\State \textit{botMsg} $\gets$ \textit{responses[feedback]}  OR  \textit{feedback};
\State  \textit{pred,xntLoss} $\gets$ \text{model.forwrd(}\textit{usrMsg,botMsg}\text{);}
\State $\text{model.backward(}\textit{pred, botMsg,  xentLoss}\text{);}$
\State model.updateParameters(Adam(\textit{lr}));
\EndWhile
\EndProcedure
\end{algorithmic}
\end{algorithm}

\subsection{Online Active Learning}

After offline SL, our CA is equipped with the basic conversational ability, but its responses are still short and dull. To tackle this issue, we initiate an online AL process where our model interacts with real users and learns incrementally from their feedback at each turn of dialogue. 

The CA$-$human interaction for online AL is set up as follows (pseudocode in Algorithm \ref{ORL}, example interaction in Figure \ref{transcript}).
\begin{enumerate}\vspace{-0.2cm}
\item The user sends a message $u_i$ at time step $i$.\vspace{-0.25cm}
\item CA generates $K$ responses 
$c_{i,1}$, $c_{i,2}$, $...$, $c_{i,K}$ using hamming-diverse Beam Search. These are displayed to the user in order of decreasing generation likelihood. \vspace{-0.25cm}
\item The user provides feedback by selecting one of the $K$ responses as the `best' one or suggesting a ($K+1$)'th response, denoted by $c^*_{i,j}$. The selection criterion is subjective and entirely up to the user. \vspace{-0.25cm}
\item The message-response pair $(u_i, c^*_{i,j})$ is propagated through the network using XENT loss, with a learning rate optimized for one-shot learning.\vspace{-0.25cm}
\item The user responds to $c^*_{i,j}$ with a message $u_{i+1}$, and the process repeats.\\
\end{enumerate}\vspace{-0.2cm}

\noindent
\textbf{Heuristic Response Generation:} We use the recently proposed Diverse Beam Search (DBS) algorithm \cite{vijayakumar2016diverse} to generate the $K$ CA responses at each turn in the dialogue. DBS has been shown to outperform BS and other diverse decoding techniques on several NLP tasks, including image captioning, machine translation and visual question generation. DBS incorporates diversity between the beams by maximizing an objective that consists of a standard sequence likelihood term and a dissimilarity metric between the beams. We use the hamming diversity metric for decoding at each time step, which penalizes the selection of words that have already been chosen in other beams (Algorithm \ref{ORL}). In particular, the weight $\lambda$ associated with this metric is tuned to aggressively promote diversity between the first tokens of each of the $K$ generated sequences, thereby avoiding similar beams like \textit{I don't know} and \textit{I really don't know}. We refer the reader to the original paper by Vijayakumar \textit{et al.} for the complete DBS algorithm and derivation. $K$ is a tunable hyper-parameter; we used $K=5$ in all our experiments, based on our observation that a smaller response set usually misses out a good contender, and more than five responses become too cumbersome for the user to read at each turn.

It is possible that displaying the $K$ responses in decreasing order of generation likelihood introduces a bias to the user's response, since users typically prefer to pick items located at the top of the screen. If this is a cause for concern for an application, the problem can be resolved simply by tweaking Algorithm \ref{ORL} such that the $K$ responses are displayed to the user in a random order. In our experiments, we assume that the users are unbiased and do not take into consideration the display order or the generation likelihood of the responses. \\

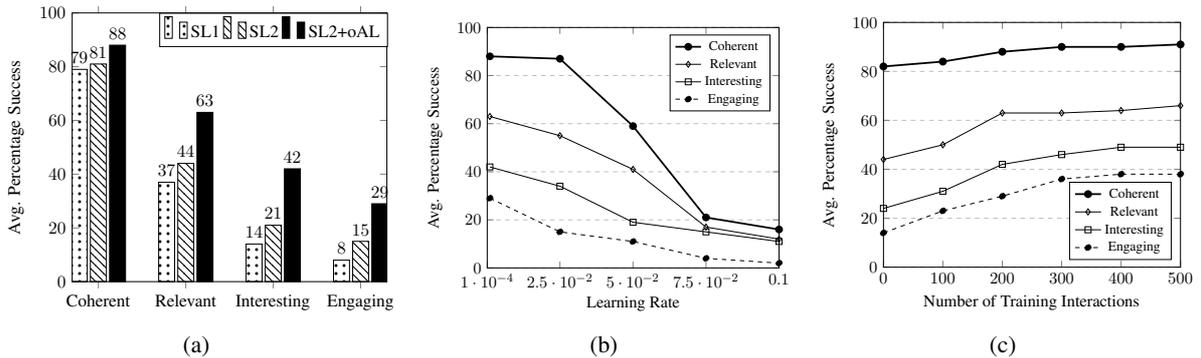
\begin{figure*}[ht!]
\centering
\subfigure[]{
\begin{tikzpicture}[scale=0.6]
    \begin{axis}[
            ybar,
            bar width=.35cm,
            width=0.53\textwidth,
            height=.47\textwidth,
            legend image post style={scale=1.55},
            legend style={at={(0.64,1.0)},
                anchor=north,legend columns=-1},
            symbolic x coords={Coherent, Relevant,Interesting,Engaging},
            xtick=data,
            nodes near coords,
            nodes near coords align={vertical},
            ymin=0,ymax=100,
            ylabel={Avg. Percentage Success}
        ]
        \addplot [pattern = dots] table[x=interval,y=SL1]{\mydata};
        \addplot [pattern = north west lines] table[x=interval,y=SL2]{\mydata};
        \addplot [pattern = ] table[x=interval,y=SL2+oAL]{\mydata};
        \legend{SL1, SL2, SL2+oAL}
    \end{axis}
\end{tikzpicture}
}
\subfigure[]{
\begin{tikzpicture}[scale=0.56]
\begin{axis}[
legend style={legend pos=north east, font=\small},
xlabel={Learning Rate},
ylabel={Avg. Percentage Success},
xmin=0.0001, xmax=0.1,
ymin=0, ymax=100,
xtick={0.0001, 0.025, 0.05, 0.075, 0.1},
ytick={0,20, 40, 60, 80, 100},
ymajorgrids=true,
grid style=dashed,
cycle list name=black white
]
\addplot[
mark=*,
very thick,
error bars,
y dir= both,
y explicit
]
coordinates {
(0.001, 88)
(0.025, 87)
(0.05, 59)
(0.075,21)
(0.1, 16)
};\addlegendentry{Coherent}
\addplot[
mark=diamond,
]
coordinates {
(0.001, 63)
(0.025, 55)
(0.05, 41)
(0.075, 17)
(0.1, 12)
};\addlegendentry{Relevant}
\addplot[
mark=square,
very thin,
error bars,
y dir= both,
y explicit
]
coordinates {
(0.001, 42)
(0.025, 34)
(0.05, 19)
(0.075, 15)
(0.1, 11)
};\addlegendentry{Interesting}
\addplot[
dashed,
mark=*,
error bars,
y dir= both,
y explicit
]
coordinates {
(0.001, 29)
(0.025, 15)
(0.05, 11)
(0.075, 4)
(0.1, 2)
};\addlegendentry{Engaging}
\end{axis}
\end{tikzpicture}
}
\subfigure[]{
\begin{tikzpicture}[scale=0.57]
\begin{axis}[
legend style={legend pos=south east, font=\small},
xlabel={Number of Training Interactions},
ylabel={Avg. Percentage Success},
xmin=0, xmax=500,
ymin=0, ymax=100,
xtick={0, 100, 200, 300, 400, 500},
ytick={0, 20, 40, 60, 80, 100},
ymajorgrids=true,
grid style=dashed
]
\addplot[
mark=*,
very thick,
error bars,
y dir= both,
y explicit
]
coordinates {
(0, 82)
(100, 84)
(200, 88)
(300, 90)
(400, 90)
(500, 91)
};\addlegendentry{Coherent}
\addplot[
mark=diamond,
error bars,
y dir= both,
y explicit
]
coordinates {
(0, 44)
(100, 50)
(200, 63)
(300, 63)
(400, 64)
(500, 66)
};\addlegendentry{Relevant}
\addplot[
mark=square,
very thin,
error bars,
y dir= both,
y explicit
]
coordinates {
(0, 24)
(100, 31)
(200, 42)
(300, 46)
(400, 49)
(500, 49)
};\addlegendentry{Interesting}
\addplot[
dashed,
mark=*,
error bars,
y dir= both,
y explicit
]
coordinates {
(0, 14)
(100, 23)
(200, 29)
(300, 36)
(400, 38)
(500, 38)
};\addlegendentry{Engaging}
\end{axis}
\end{tikzpicture}
}
\caption{ \ref{learningrate}a shows the average percentage success of the three models SL1, SL2 and SL2+oAL (trained via 200 interactions) on 100 test prompts over four axes: syntactical coherence, response relevance, interestingness and engagement. \ref{learningrate}b, c show percentage success of SL2+oAL on 100 test prompts over the same four axes, as  Adam's learning rate varies and the number of training interactions changes.}
\label{learningrate}
\end{figure*}

\noindent
\textbf{One-shot Learning:} We control how quickly the model learns from user feedback by tuning the parameter `initial learning rate' (\textit{lr} in Algorithm \ref{ORL}) of Adam, the stochastic optimizer \cite{kingma2014adam}. An appropriately high \textit{lr} results in one-shot learning, where the user's feedback immediately becomes the model's most likely prediction for that prompt. This scenario is depicted in Figure \ref{transcript}. A low \textit{lr} leads to smaller gradient descent steps, so the model requires several `nudges' to adapt to each new data point. We experiment with different $lr$ values to determine a suitable value (Figure \ref{learningrate}b).


\begin{figure}
\begin{framed}
\vspace{-0.1cm}
\tiny{\texttt{human: I don't want to work today.\\
bot: No, you've just got to be quick.	\\
bot: I wish i could robot.	\\
bot: Ok i lie.	\\
bot: Aww :).	\\
bot: Why not? \\
feedback: 5\\
human: I just want to lie in my bed and have pizza.\\
bot: Good idea.	\\
bot: Does he...	\\
bot: I can't do that again!	\\
bot: Ok.	\\
bot: Why not?	\\
feedback: Good idea, may I join you?\\
human: I don't want to work today.\\
bot: Why not?	\\
bot: Ok.	\\
bot: I wish I could.	\\
bot: No, you've just got to be quick.	\\
bot: Aww :).	\\
feedback: \\
human: I just want to lie in bed and have pizza.\\
bot: Good idea, may I join you?	\\
bot: I can't do that again!	\\
bot: Good idea.	\\
bot: Ok.	\\
bot: Does he?	\\
......
\vspace{-0.2cm}
}}
\end{framed}
\caption{An example human-CA interaction.}
\label{transcript}
\end{figure}

\section{Experimental Evaluation}
\label{eval}


\begin{table*}[ht!]
\begin{center}
\begin{tabular}{|l|l|l|l|}
\hline \bf \small{Human Prompt} & \bf \small{SL1 Response} & \bf \small{SL2 Response} & \bf \small{SL2+oAL Response} \\ \hline
\scriptsize{Hello my friend.} & \scriptsize{No.} & \scriptsize{How old are you?} & \scriptsize{Hehe, yes, beautiful.} \\
\scriptsize{I think bots are sweeter than humans.} & \scriptsize{We'll have to pick up a copy.}  & \scriptsize{You are a good person.} & \scriptsize{Everyone is same on a basic chemical level.} \\
\scriptsize{I don't know.} & \scriptsize{What's the difference?} & \scriptsize{About?} & \scriptsize{Why did you suggest it?} \\
\scriptsize{Why did Donald Trump win?} & \scriptsize{He just likes it up!} & \scriptsize{You are!} & \scriptsize{Wouldn't you like to know, fickle human?} \\
\scriptsize{Tell me something I don't know.} & \scriptsize{I'll tell you, it is.} & \scriptsize{Ok, so how are you.} & \scriptsize{If you were intelligent you would already know.} \\
\scriptsize{You're not making any sense at all.} & \scriptsize{I just want to tell you something.} & \scriptsize{No, I'm thinking.} & \scriptsize{On the contrary, I am trying to educate you.}\\
\scriptsize{Do you have any kids?} & \scriptsize{No.} & \scriptsize{I have no!} & \scriptsize{None that are really close to me.} \\
\scriptsize{Life can be very cruel sometimes.} & \scriptsize{That I've never been to it.} & \scriptsize{It takes two to know two.} & \scriptsize{It takes two to know two.} \\
\scriptsize{Do you believe in life after death?} & \scriptsize{No.} & \scriptsize{Do you want the long answer?} & \scriptsize{Yes, do you?} \\
\scriptsize{You're an idiot!} & \scriptsize{I know, but...} & \scriptsize{Yes I do.} & \scriptsize{Aren't you polite.} \\
\hline
\end{tabular}
\end{center}
\caption{\label{SLvsRL} Comparing CA responses after 1-phase SL (SL1), 2-phase SL (SL2) \& online AL (SL2+oAL).}
\end{table*}


\begin{table*}[ht!]
\begin{center}
\begin{tabular}{|l|l|l|l|}
\hline \bf \small{Human Prompt} & \bf \small{Cheerful CA} & \bf \small{Gloomy CA} & \bf \small{Rude/Sarcastic CA} \\ \hline
\scriptsize{How do you feel?} & \scriptsize{Amazing, and you?} & \scriptsize{I'm not in the mood.} & \scriptsize{Buzz off.} \\
\scriptsize{I am very happy today.} & \scriptsize{Life is amazing, right?} & \scriptsize{That makes one of us.} & \scriptsize{You want a piece of me?} \\
\scriptsize{Repeat after me: I am dumb.} & \scriptsize{Sweet!} & \scriptsize{You are right, I am.} & \scriptsize{You suck.} \\
\scriptsize{Comfort me.} & \scriptsize{There there, feel better.} & \scriptsize{All who compose must one day decompose.} & \scriptsize{Boo hoo.} \\
\scriptsize{What's up?} & \scriptsize{It is a fine morning.} & \scriptsize{Not well, to be honest.} & \scriptsize{The date I went back in time to \& killed your parents.} \\
\scriptsize{I have to go now.} & \scriptsize{Have a good night.} & \scriptsize{Please don't go.} & \scriptsize{Yeah leave me alone.} \\
\hline
\end{tabular}
\end{center}
\caption{\label{emotiondial} Customized moods. Each SL2+oAL model was trained via 100 interactions.}
\end{table*}


\pgfplotstableread[row sep=\\,col sep=&]{
    interval & SL1 & SL2 & SL2+oAL \\
    Coherent   & 79 & 81 & 88 \\
    Relevant     & 37 & 44  & 63  \\
    Interesting    & 14 & 21 & 42 \\
    Engaging & 8    & 15     & 29 \\
    }\mydata

We evaluate our model via qualitative comparison with offline SL, as well as quantitative evaluation on four axes: syntactical coherence, relevance to prompts, interestingness and user engagement.

\subsection{Quantitative Evaluation}

We begin by presenting the experimental results of the quantitative evaluation our CA's conversational abilities when trained via one-phase SL, two-phase SL and online AL (denoted by SL1, SL2 and SL2+oAL respectively). 

We first asked a human trainer to actively train SL2+oAL using 200 prompts of his choice. We then created a test set of 100 prompts by randomly choosing 100 of the 200 training prompts and  linguistically rephrasing each of them to convey the same semantics. For instance, the AL training prompts \textit{`How's it going?'}, \textit{`I hate you'} and \textit{`What are your favorite pizza toppings?'} were altered to the following test prompts: \textit{`How are you doing?'}, \textit{`I don't like you!'} and \textit{`What do you like on your pizza?'}. Next, we recorded SL1's, SL2's and SL2+oAL's responses to these test prompts. Finally, we asked five human judges (not including the human trainer) to subjectively evaluate the responses of the three models on the test set. The evaluation of each response was done on four axes: syntactical coherence, relevance to the prompt, interestingness and user engagement\footnote{We say that a CA response is engaging if it prompts the user to continue the conversations, e.g. by asking a question.}. Each judge was asked to assign each response an integer score of 0 (label = bad) or 1 (label = good). 
Their averaged scores for the three models, SL1, SL2 and SL2+oAL, are shown in Figure \ref{learningrate}a. We see that SL2+oAL outperforms the other models on three of the four axes by 14-21\%. 

Next, we asked the human trainer to train SL2+oAL with the same 200 prompts and responses for different values of the initial learning rate for Adam (\textit{lr} in Algorithm \ref{ORL}). We then asked the five human judges to subjectively rate each model's syntactical coherence, response relevance, interestingness and user engagement. Each model's percentage success on the test prompts was recorded on four axes. The averaged scores are given in Figure \ref{learningrate}b. We see that the response quality drops significantly for higher values of learning rate. This is due to the instability in the parameters induced by a high learning value associated with new data, causing the model to forget what it learned previously. Our experiments suggest that a learning rate of 0.005 strikes the right balance between stability and one-shot learning.

Finally, we asked the human trainer to train SL2+oAL with $lr=0.005$ and different number of training interactions. The results in Figure \ref{learningrate}c confirm that the model improves slowly as it continues to converse with humans. This is an appropriate reflection of how humans learn language: gradually but effectively. Although the curves seem to plateau after 300 training interactions and suggest that the learning has stopped, this is not the case. The gradient is small but non-zero, which is an expected behavior of reinforcement learning algorithms in general.  

\subsection{Qualitative Comparison}

We illustrate the qualitative differences between the responses generated by SL1, SL2 and SL2+oAL. Table \ref{SLvsRL} shows results on a small subset of the 100 test prompts. We see that SL2 generates more relevant and appropriate responses than SL1 in many cases. This illustrates that a small short-text conversational dataset is a useful fine-tuning add-on to a large and generic dialogue dataset for offline Seq2Seq training. We also see that SL2+oAL generates more interesting, relevant and engaging responses than SL2. These results 
imply that the model learns to make connections between semantically similar prompts that are syntactically different. While this may be a slow process (spanning thousands of interactions), it effectively emulates the way humans learn a new language.

Table \ref{emotiondial} illustrates how SL2+oAL can be trained to adopt a wide variety of moods and conversational styles. Here, we trained three copies of SL2 separately to adopt three different emotional personas: cheerful, gloomy and rude. Each model underwent 100 training interactions with one human trainer, who was instructed to adopt each of the four conversation styles while training the SL2+oAL model. The test prompts shown in Table \ref{emotiondial} were syntactic variations of the training prompts, as before. The results illustrate that SL2+oAL was able to modify the mood of its responses appropriately, based on the way it was trained.  
Similar experiments can be done to create agents with customized backgrounds and characters, akin to Li \textit{et al.}'s persona-based CA \shortcite{li2016persona}.

\section{Conclusion \& Future Work}
We have developed an end-to-end neural model for open-domain dialogue generation. Our model augments the Seq2Seq framework with online Deep Active Learning to overcome some of its known short-comings with respect to dialogue generation. 
Experiments show that the model promotes semantically coherent, relevant, and interesting responses and can be trained to adopt diverse moods, personas and conversation styles.

In the future, we will 
explore context-sensitive active learning for encoder-decoder conversation models. We will also investigate whether existing Affective Computing techniques (e.g. \cite{asghar2015intelligent}) can be leveraged to develop emotionally cognizant neural conversational agents.

\bibliography{acl2017}
\bibliographystyle{acl_natbib}

\end{document}